\documentclass[10pt,twocolumn,letterpaper]{article}

\usepackage{iccv}
\usepackage{times}
\usepackage{epsfig}
\usepackage{graphicx}
\usepackage{amsmath}
\usepackage{amssymb}

\usepackage{multirow}
\usepackage{booktabs}
\usepackage{xcolor}
\usepackage{enumitem}
\usepackage{pdfpages}

\usepackage[numbers,sort,compress]{natbib}

\definecolor{darkgreen}{rgb}{0,0.5,0}
\definecolor{frenchblue}{rgb}{0.0, 0.45, 0.73}

\usepackage[pagebackref=true,breaklinks=true,letterpaper=true,colorlinks,bookmarks=false]{hyperref}

\iccvfinalcopy 

\def\iccvPaperID{3411} 

\ificcvfinal\fi

\begin{document}

\title{Render In-between: Motion Guided Video Synthesis for Action Interpolation
}

\author{Hsuan-I Ho$^{1}$
\and
Xu Chen$^{1,2}$
\and
Jie Song$^{1}$
\and
Otmar Hilliges$^{1}$
\and
$^{1}$Department of Computer Science, ETH Zürich \\
$^{2}$Max Planck Institute for Intelligent Systems, Tübingen
}

\maketitle
\ificcvfinal\thispagestyle{empty}\fi

\begin{abstract}
Frame rate greatly impacts the style and viewing experience of a video. Especially for footage that depicts fast motion, higher frame rate equates to smoother imagery and reduced motion blur artifacts. Upsampling of low-frame-rate videos of human activity is an interesting yet challenging task with many potential applications ranging from gaming to entertainment and sports broadcasting.   
The main difficulty in synthesizing video frames in this setting stems from the highly complex and non-linear nature of human motion and the complex appearance and texture of the body. 
We propose to address these issues in a motion-guided frame-upsampling framework that is capable of producing realistic human motion and appearance. 
A novel motion model is trained to interpolate the non-linear skeletal motion between frames by leveraging a large-scale motion-capture dataset (AMASS). The high-frame-rate pose predictions are then used by a neural rendering pipeline to produce the full-frame output, taking the pose and background consistency into consideration. 
Our pipeline only requires low-frame-rate videos and unpaired human motion data but does not require high-frame-rate videos for training. 
Furthermore, we contribute the first evaluation dataset that consists of high-quality and high-frame-rate videos of human activities for this task. 
Compared with state-of-the-art video interpolation techniques, our method produces interpolated frames with better quality and accuracy, which is evident by state-of-the-art results on pixel-level, distributional metrics and comparative user evaluations. Our code and the collected dataset are available at \url{https://git.io/Render-In-Between}.
\end{abstract}

\section{Introduction}

\begin{figure}[t]
    \includegraphics[width=\linewidth]{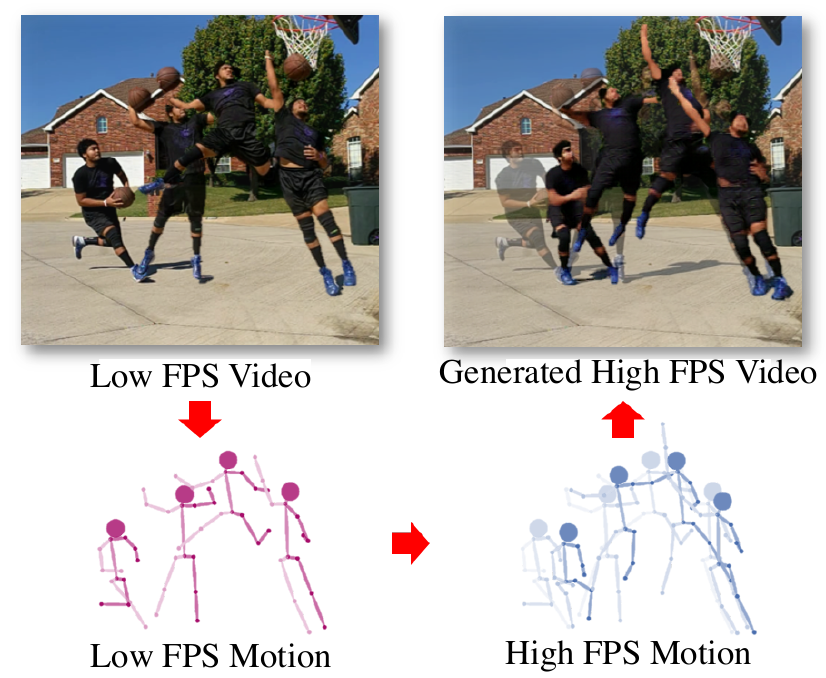}
    \caption{\textbf{Motion guided video interpolation.} Given a low-frame-rate video of human activities, our method produces a high-frame-rate video by explicitly modeling human motion, guiding a neural renderer. We first extract 2D pose sequences from the video, then non-linearly interpolate the pose sequence via a learned human motion model, and finally generate images that correspond to the high-frame-rate pose sequence. }
    \label{fig:teaser}
\end{figure}



High-frame-rate videos of human activity have many applications. For example in sports or multimedia production, but
also in user input sensing~\cite{Xu_2020_CVPR,kowdle2018need} and as a source of training data for discriminative tasks such as activity recognition. 
However, most existing video material is captured at low frame rates, typically at 24 or 30 frames per second (FPS), and capturing higher FPS video requires special equipment, good lighting conditions, and increases storage/bandwidth needs. 
In this paper, we propose a novel architecture for the temporal upsampling of low FPS videos that are capable of generating videos of high visual fidelity by explicitly reasoning about human motion to guide a neural renderer.


Recently, some attempts have been made to synthesize smooth slowmotion clips from videos recorded at standard frame rates. Video frame interpolation techniques aim at generating intermediate frame(s) to produce a video sequence at higher frame rates. Conventionally this is solved by estimating optical flow~\cite{baker2011database,jiang2018super,bao2019depth}, which depicts the per-pixel motion between consecutive frames, and by warping input frames accordingly. Powered by recent advances in deep learning, a newly emerging line of work attempts to generate intermediate frames directly from input frames. Such methods bypass the explicit optical flow estimation but infer and utilize the motion implicitly via a learned convolution kernel~\cite{Niklaus_CVPR_2017,Niklaus_ICCV_2017} or feature shuffling~\cite{choi2020cain}.

However, in the presence of the complicated nature of human motion, both approaches face severe challenges. Since optical flow assumes small local motion, methods based on optical flow struggle to generate video frames containing fast motion and large displacements.
CNN based methods typically struggle in representing motion with displacements that exceed their receptive field.
Furthermore, both of the above approaches either implicitly or explicitly assume linear motion and hence have difficulties with the non-linear dynamics and complex
articulation patterns of human motion. As a result, artifacts such as ghosting limbs and blurry bodies are frequent artifacts in temporal upsampling of videos depicting human activities (See Fig.~\ref{fig:comparison}).


To address the challenges caused by the inherent properties of human motion, we propose a new pipeline to tackle this problem. At the heart of our approach lies the concept of leveraging an explicit model of human motion to better capture the dynamics of the foreground motion. The proposed motion model is trained to predict plausible human motion at a high frame rate. These joint predictions are then used to guide a neural rendering model to generate the corresponding human images. Compared to end-to-end video interpolation models, our learning-based motion model can better capture non-linear motion details than optical flow, and the neural rendering model can benefit from this conditional image generation scheme to synthesize realistic human bodies and cloth textures ~\cite{wang2018pix2pixHD}.

More specifically, given a low-frame-rate video of human activities, we first extract 2D skeletons with a 2D pose estimator \cite{fang2017rmpe}. Next, a transformer-based motion model predicts joint positions for the intermediate time steps. The motion model is trained on a large-scale MoCap dataset (AMASS~\cite{AMASS:ICCV:2019}) and thus can predict plausible joint trajectories for a wide range of activities. 
It performs the dual task of i) reducing possible detection errors and missing joints from the input body poses, and ii) predicting plausible joint trajectories for a wide range of activities at a high frame rate. 
Finally, a neural rendering model generates a sequence of foreground human bodies corresponding to the 2D joint trajectories. To further take the background dynamics and consistency into consideration, our model predicts alpha blending masks to composite generated foreground bodies with warping based background images (e.g.~\cite{bao2019depth,xu2019quadratic}), yielding the final output sequences at high frame rates. It is worth noting that our method requires only low-frame-rate videos and a separate MoCap dataset as training data. No high-frame-rate ground truth is required for supervision.


Furthermore, we contribute a new dataset of high-frame-rate and high-quality human activity videos for quantitative and qualitative evaluation purposes. The experimental results indicate that such videos are indeed challenging for the existing video interpolation techniques and that our method yields better results evident by the pixel-level, distributional metrics and user studies. In further ablations, we verify the effectiveness of our design choices in both modules. In summary, we make the following contributions:


\begin{itemize}[nosep]
    \item A novel video frame-upsampling framework for human activities by combining,
    \begin{itemize}[nosep]
         \item a non-linear motion modeling network that is robust to occlusion and incomplete input 
          \item a human image generation scheme that handles complex human geometry and dynamic background scenes.
    \end{itemize}
    \item A new dataset of high-quality and high-speed human activities for evaluation.
\end{itemize}

\section{Related Works}
\begin{figure*}[t]
\begin{minipage}[]{1.0\linewidth}
  \centering
  \centerline{\includegraphics[width=1.0\linewidth]{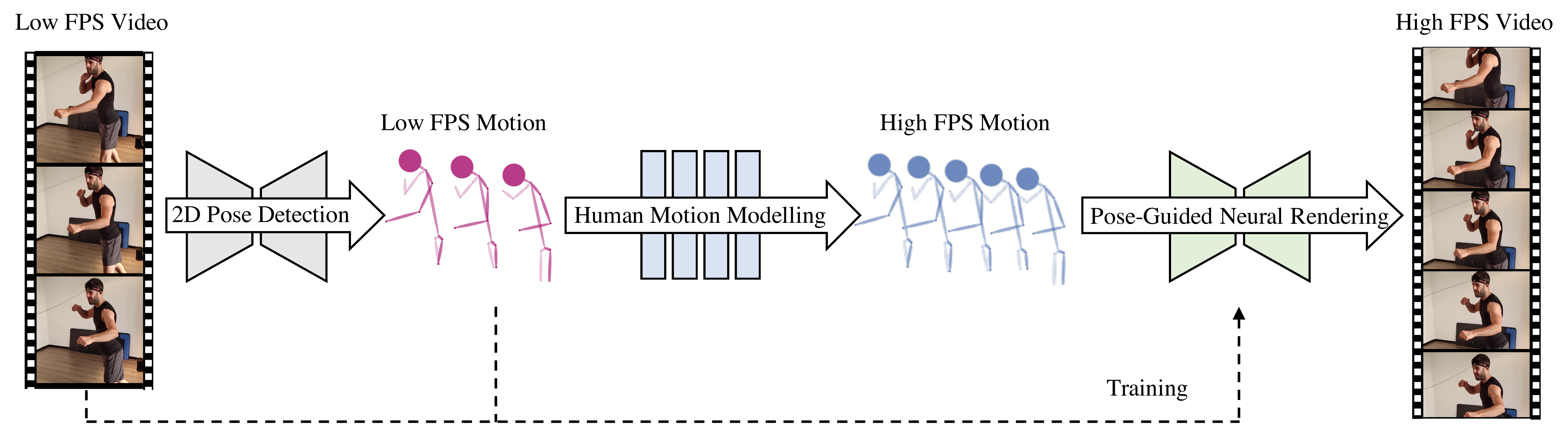}}
\end{minipage}
\vspace{2pt}
\caption{\textbf{Method overview.} Given a low FPS video as input, we first extract the 2D keypoints of each frame using off-the-shelf 2D pose detector \cite{fang2017rmpe}, obtaining a low FPS pose sequence. Based on this low FPS sequence, we infer 2D poses at intermediate time steps via a learned human motion model to obtain a high FPS pose sequence. Finally, guided by 2D poses in the high FPS sequence, images at intermediate time steps are generated using a neural rendering model to form the desired high FPS video. This neural rendering model is trained specifically on the low FPS video frames and the corresponding poses only, without requiring high FPS videos as training data.}
\label{fig:framework}
\end{figure*}

\paragraph{Video Frame Interpolation.}
Conventional approaches \cite{herbst2009occlusion,baker2011database,werlberger2011optical} to video interpolation often utilize optical flow. Pixels in the input frames are warped along the optical flow direction to produce the intermediate frames. Deep learning techniques have also been integrated recently to predict a 3D optical flow over space and time for warping input frames~\cite{liu2017video}, or to fuse warped frames~\cite{jiang2018super, bao2019depth,Niklaus_CVPR_2020,choi2020deep,xu2019quadratic, BMBC, chi2020all}. Powered by recent advances in deep learning, a newly emerging line of work \cite{Niklaus_CVPR_2017,Niklaus_ICCV_2017,liu2019deep,choi2020cain} attempts to generate intermediate frames directly from input frames. For instance, Long \etal~\cite{long2016learning} leverage a CNN to predict the intermediate frame between two consecutive frames. Niklaus \etal~\cite{Niklaus_CVPR_2017,Niklaus_ICCV_2017} learn to predict a kernel to fuse input frames by convolution. Choi \etal~\cite{choi2020cain} shuffle the features of two images with PixelShuffle~\cite{shi2016real} and then fuse them with an attention-based image generator. 
However, videos of high-speed human activities are challenging for the above approaches. As noted above, these general-purpose methods would produce artifacts under fast non-linear human body motion. This is why a specialized framework (as ours) is practically preferable, with the goal of producing more realistic body motion and less artifacts at high frame rate.

\paragraph{Human Image Generation.}
The computer graphics literature has dedicated much attention to this problem, including skinning and articulating 3D meshes, simulation of physically accurate clothing deformation, and the associated rendering problems~\cite{guan20102d,chen2016synthesizing,goldenthal2007efficient,pishchulin2012articulated,narain2012adaptive,kulkarni2015deep}. Despite much progress, generating photo-realistic renderings remains difficult and is computationally expensive. 

With the advances of deep image generation networks, methods have been proposed to generate human images with diverse appearances, clothing, and poses by learning from data \cite{Lassner:GeneratingPeople:2017,NIPS2017_6644,han2017viton,wang2018toward,raj2018swapnet,zhao2018multi,zanfir2018human,esser2018variational,siarohin2018deformable,balakrishnan2018synthesizing,pumarola2018unsupervised,dong2018soft,grigorev2019coordinate,song2019unsupervised,liu2019liquid,wang2019few,chan2019dance,liu2019neural,han2019clothflow,men2020controllable,Gafni2020Vid2Game,sun2020human,ma2020unselfie,ren2020deep,weng2020misc,transmomo2020}. The problem has been typically addressed by leveraging conditional image generation techniques \cite{wang2018pix2pixHD} to map human body representations, such as 2D skeleton or a projection of 3D human body meshes, to realistic human images. Among these works, human reposing approaches are especially related to this paper. Given an image or a video of a person, such approaches aim to generate images of the person in desired poses. The person's identity is preserved via image or feature warping operations \cite{siarohin2018deformable,balakrishnan2018synthesizing,liu2019liquid,ren2020deep}, few-shot adaption \cite{wang2019few}, or person-specific networks \cite{yang2018pose, cai2018deep,chan2019dance,wang2018vid2vid,liu2019neural,Gafni2020Vid2Game,sun2020human}. 
While this human reposing task assumes natural body pose as input, the task of video interpolation further requires generating realistic body poses, which we address in this paper. 

\paragraph{Human Motion Modelling.}
Modelling 3D human motion is an important task in computer graphics and vision. Given an input motion sequence, which is typically represented by the positions and orientations of body joints, methods have been proposed to predict future motion~\cite{jain2016structural,ghosh2017learning,martinez2017human,li2018convolutional,gui2018adversarial,zhou2018auto,mao2019learning,hernandez2019human,li2018convolutional,Aksan_2019_ICCV,aksan2020attention}, change the style of the motion~\cite{Villegas_2018_CVPR,aberman2019learning, transmomo2020} or infill intermediate motion~\cite{harvey2018recurrent,harvey2020robust,kaufmann2020infilling}. 
These methods all assume clean 3D body poses as inputs. In our case, however, only 2D poses can be extracted from the video. Thus the problem becomes more challenging since we need to handle the large variety of scale and translation of the skeleton caused by the perspective projection. We use a large-scale motion dataset to simulate 2D poses as observed by a pinhole camera to train a 2D motion interpolation network. Moreover, to handle noisy 2D joint data, we further augment the simulated samples by random perturbations and dropouts of joints.


\section{Method}
\label{sect:method}

Given a low-frame-rate (low FPS) video $V_{low}\!\!=\!\!\{ I_0,I_1,...,I_{T}\}$ of human activities consisting of $T\!\!\!+\!\!\!1$ frames, our goal is to generate a high FPS video $\hat{V}_{high}\!\!\!=\!\!\!\{ \hat{I}_0,\hat{I}_{1/s},\hat{I}_{2/s},...,\hat{I}_1,...,\hat{I}_{T}\}$ consisting of $sT\!\!+\!\!1$ frames. 

The focus of this paper is to tackle the challenging case of human activities. Our key idea is to explicitly model human motion to guide the generation of intermediate frames in order to capture the non-linear dynamics and complex articulation patterns of human motion. Following this idea, we propose a two-stage pipeline, as illustrated in Fig.~\ref{fig:framework}, combining learning-based human motion modeling and pose-guided neural rendering.

\subsection{Overview}
The input to our pipeline is a low FPS video. For each frame $I_t$ in the given low FPS video, we first use an off-the-shelf 2D pose estimation model~\cite{fang2017rmpe} to extract 2D human skeletons
$\boldsymbol{\bar{p}}_t \in \mathbb{R}^{2 \times J}$,
which is represented by 2D positions of $J=19$ body joints.
With the low FPS pose sequence 
$\boldsymbol{\bar{P}}_{low}= \{\boldsymbol{\bar{p}}_0,\boldsymbol{\bar{p}}_1,...,\boldsymbol{\bar{p}}_T \}$,
we then determine the pose $\boldsymbol{\hat{p}}_{\tau}$ at intermediate time steps $\tau \in [0,T]$ using a learned human motion model to form the high FPS sequence 
$\boldsymbol{\hat{P}}_{high}= \{\boldsymbol{\hat{p}}_0,\boldsymbol{\hat{p}}_{1/s},\boldsymbol{\hat{p}}_{2/s},...,\boldsymbol{\hat{p}}_1,...\boldsymbol{\hat{p}}_T \}$. Details on network architecture and training scheme in Sec.~\ref{sect:motion}.

Subsequently, we train a pose-guided neural rendering model specifically for the given video using the low FPS frames $V_{low}$ and the corresponding poses $~\boldsymbol{\bar{P}}_{low}$. The trained model can map body poses $\boldsymbol{p}$ to human images $I$. We feed each high FPS pose sample $\boldsymbol{\hat{p}}_{\tau}$ obtained from the previous stage to generate the desired high FPS video $\hat{V}_{high}$. For details of neural rendering please see Sec.~\ref{sect:generation}.

\subsection{Human Motion Modelling}
\label{sect:motion}
\begin{figure}[t]
\begin{minipage}[]{1.0\linewidth}
  \centering
  \centerline{\includegraphics[width=1.0\linewidth]{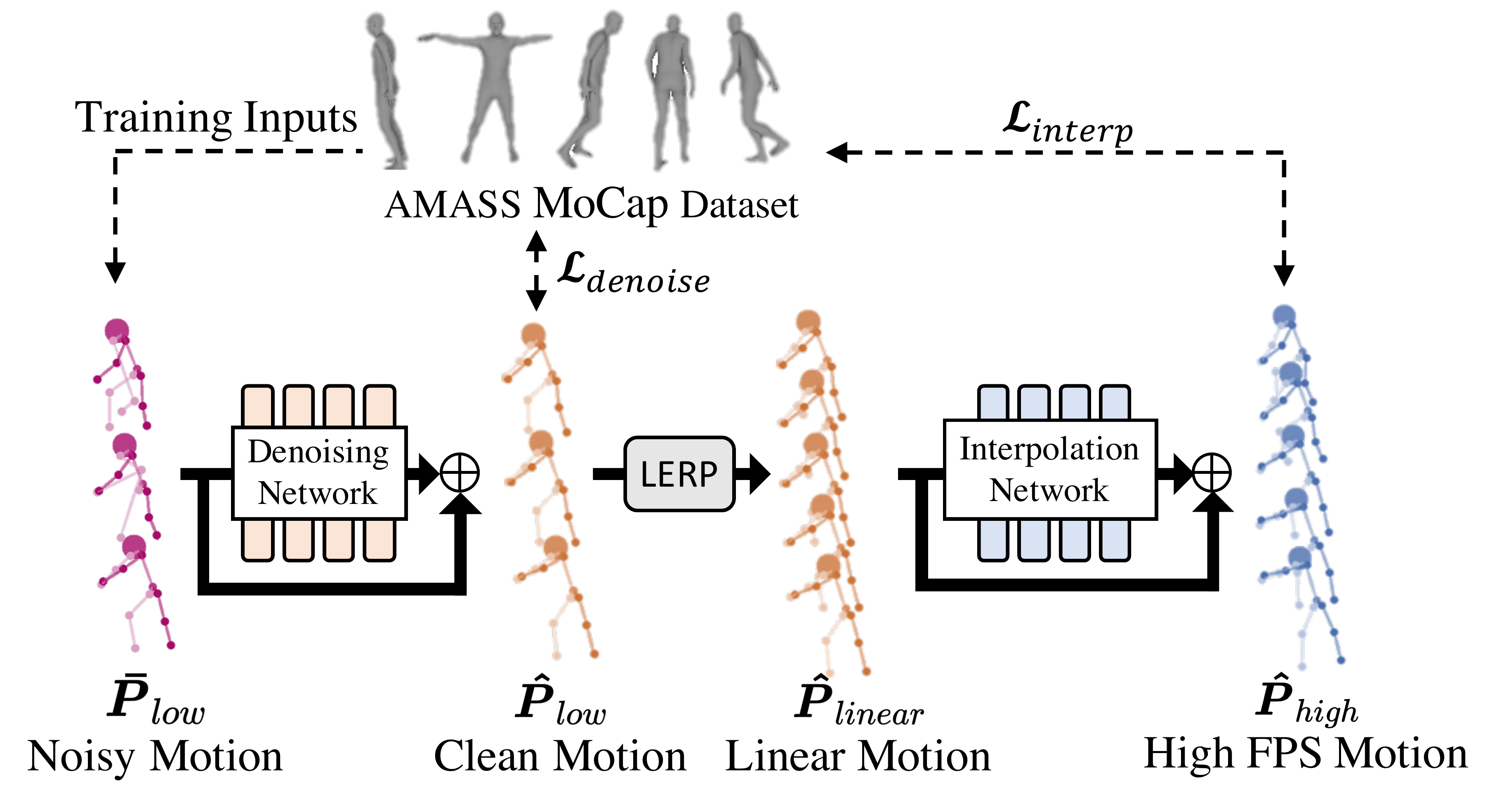}}
\end{minipage}

\caption{\textbf{Human motion model.} Given a low FPS noisy pose sequence as input (\emph{left}), our denoising network first reduces the noise from the input. The denoised sequence is then linearly interpolated (\emph{LERP}) to form the linear component of high FPS motion. Finally, the network infers the non-linear motion component, obtaining high FPS motion by summing the linear and non-linear components together. The required training data is obtained from simulating 2D keypoints using a large-scale human motion dataset AMASS.} 
\label{fig:transformer}
\end{figure}

The goal of the motion modelling stage, as illustrated in Fig.~\ref{fig:transformer}, is to generate a high FPS realistic human motion sequence $\boldsymbol{\hat{P}}_{high}$ from a noisy low FPS pose sequence $\boldsymbol{\bar{P}}_{low}$.
\paragraph{Motion Denoising Network.}
\label{sect:4.1}

A major challenge to apply learning-based human motion modelling to our case is that our input motion sequences are often noisy and even incomplete, due to the limited accuracy of 2D pose estimation or possible occlusion. This conflicts with the common assumption of noise-free motion in the existing motion modelling approaches.

We interpret the problem as a sequence-to-sequence translation task, and use a deep network $T_{denoise}$ to map the noisy input pose sequence $\boldsymbol{\bar{P}}_{low}$ to its clean version $\boldsymbol{\hat{P}}_{low}$. The network predicts a correction term which is added to the input to reduce the noise:
\begin{equation}
\begin{split}
\boldsymbol{\hat{P}}_{low}&= \boldsymbol{\bar{P}}_{low} + T_{denoise}(\boldsymbol{\bar{P}}_{low})
\end{split}
\end{equation}
\paragraph{Motion Interpolation Network.}

While linear motion models can approximate some of the joint displacements, the residual non-linear component has a significant influence on the realism of the motion.
We thus model this non-linear component via a learned deep network $T_{interp}$. 
Starting by linearly interpolating the clean but low FPS pose sequence $\boldsymbol{\hat{P}}_{low}$, a high FPS linear motion sequence $\boldsymbol{\hat{P}}_{linear}$ is treated as an initial approximation of the high FPS motion.
Given $\boldsymbol{\hat{P}}_{linear}$, we then recover the missing non-linear components using the interpolation network $T_{interp}$. The final output, i.e., the high FPS motion $\boldsymbol{\hat{P}}_{high}$, is the sum of both linear and non-linear components, given by 
\begin{equation}
\boldsymbol{\hat{P}}_{high}= \boldsymbol{\hat{P}}_{linear} + T_{interp}(\boldsymbol{\hat{P}}_{linear})
\end{equation}

\paragraph{Network Backbone.}
Motivated by recent successes in other generative tasks ~\cite{vaswani2017attention,aksan2020attention}, we adopt the transformer architecture as the backbone of both networks $T_{denoise}$ and $T_{interp}$. We experimentally verify this choice and compare it to alternatives such as RNNs and CNNs in Sec.~\ref{sec:ablation_motion}.

\paragraph{Training.}
\label{sec:motion_training}

Training our networks requires large-scale realistic 2D human motion sequences, which are scarce due to the known difficulty of annotating sequential data. Therefore, we propose to simulate the required training data via the AMASS MoCap dataset~\cite{AMASS:ICCV:2019}. This dataset contains high FPS sequences of realistic human poses and shapes parameterized by the SMPL~\cite{SMPL:2015} model.
For each high FPS motion sample, we first generate its 3D body joints from the pose and shape parameters and project these joints onto the image plane using a perspective camera, where the intrinsics and extrinsics are sampled within a expected value range that fits the image coordinate, obtaining a sequence of 2D joints $\boldsymbol{P}_{high}$.
This ground truth motion $\boldsymbol{P}_{high}$ is downsampled to generate its low FPS counterpart $\boldsymbol{P}_{low}$ for training. To further simulate detection noise and missing joints, we perturb some keypoints with random noise and randomly drop-out joints to attain the training sample $\boldsymbol{\bar{P}}_{low}$.

Given the noisy low FPS motion $\boldsymbol{\bar{P}}_{low}$ at training time, the network recovers the clean and high FPS motion $\boldsymbol{\hat{P}}_{high}$, which is used for computing loss with the ground truth $\boldsymbol{P}_{high}$.
The training loss consists of two terms, one for denoising $\mathcal{L}_{denoise}$ and the other for interpolation $\mathcal{L}_{interp}$. The denoising term penalizes the difference between the output of our denoising network and the low FPS clean motion $\boldsymbol{P}_{low}$, which is defined as:
\begin{equation}
    \mathcal{L}_{denoise} = \left \|\boldsymbol{P}_{low} -  \boldsymbol{\hat{P}}_{low} \right \|_{1}.
\end{equation} 
The interpolation term penalizes the difference between the output of our interpolation network and the high FPS motion $\boldsymbol{P}_{high}$:
\begin{equation}
    \mathcal{L}_{interp} = \left \| \boldsymbol{P}_{high} - \boldsymbol{\hat{P}}_{high} \right \|_{1}.
\end{equation}
The final loss is defined as:
\begin{equation}
    \mathcal{L}_{motion} = \mathcal{L}_{denoise} + \lambda_{interp} \mathcal{L}_{interp},
\end{equation}
where $\lambda_{interp}$ controls the weights of the two terms.


\subsection{Posed-Guided Neural Rendering}
\label{sect:generation}

\begin{figure}[t]
\begin{minipage}[]{1.0\linewidth}
  \centering
  \centerline{\includegraphics[width=1.0\linewidth]{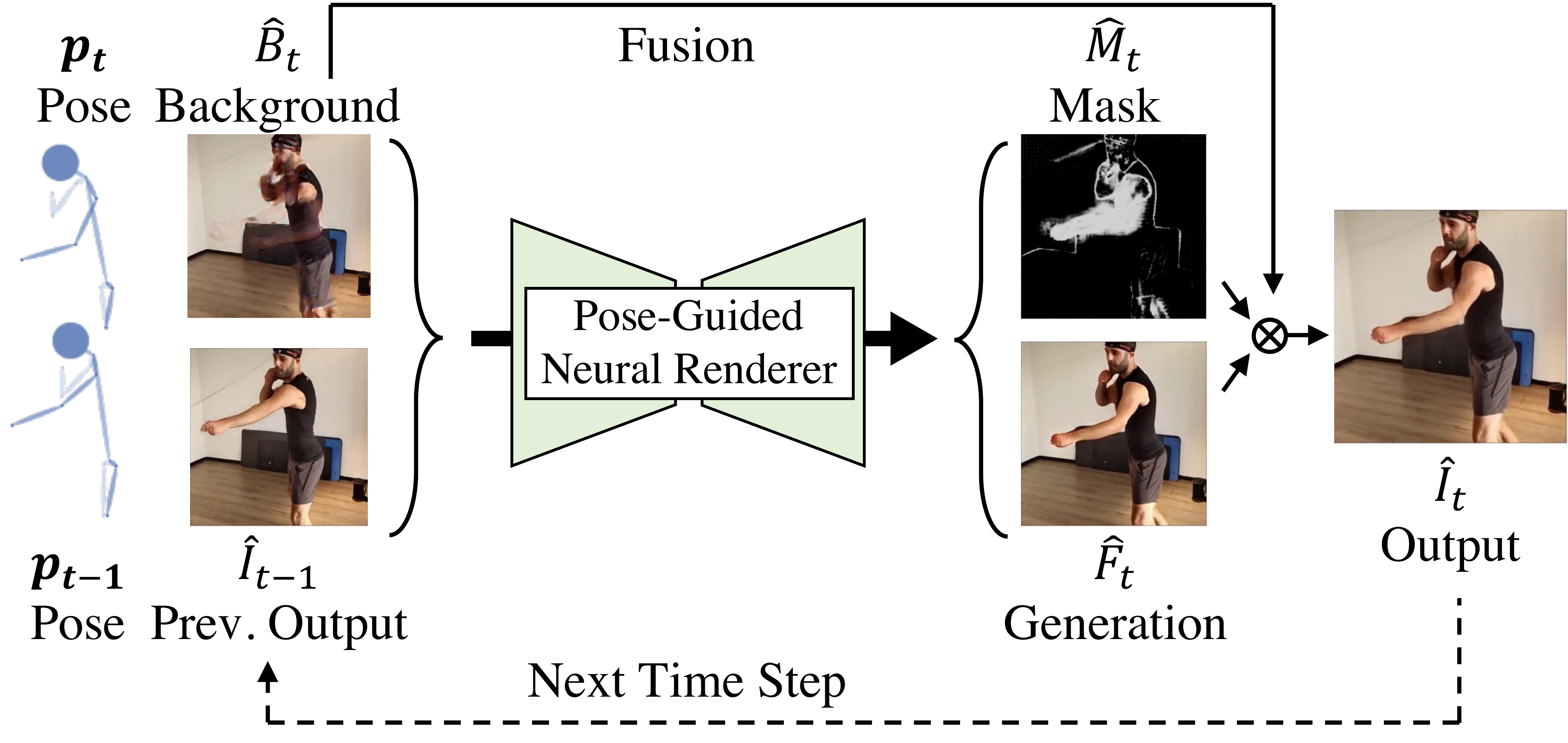}}
\end{minipage}
\caption{\textbf{Pose-guided neural rendering.} Our neural rendering model maps body poses to human images. The inputs to the network consist: an image showing 2D joints and connecting limbs to control the pose of the generated human image (\emph{pose $\boldsymbol{p}_t$}), an background image generated by off-the-shelf video interpolation model to handle background motion (\emph{background $B_t$}), the pose and the generated image at the previous time step to encourage temporal coherency (\emph{pose $\boldsymbol{p}_{t-1}$ and previ. output $\hat{I}_{t-1}$)}. The model generates a human image with the desired body pose $\hat{F}_{t}$ alongside an alpha mask $\hat{M}_{t}$ that fuses the generation $\hat{F}_{t}$ and the background $\hat{B}_{t}$ into the final output $\hat{I}_{t}$.
}
\label{fig:generation}
\end{figure}

After obtaining the high FPS pose sequence 
$\boldsymbol{\hat{P}}_{high}= \{\boldsymbol{\hat{p}}_0,\boldsymbol{\hat{p}}_{1/s},\boldsymbol{\hat{p}}_{2/s},...,\boldsymbol{\hat{p}}_1,...\boldsymbol{\hat{p}}_T \}$
we can generate the high FPS video $\hat{V}_{high}=\{ \hat{I}_0,\hat{I}_{1/s},\hat{I}_{2/s},...,\hat{I}_1,...,\hat{I}_{T}\}$ via a learned neural rendering model $G: \boldsymbol{p} \rightarrow I$. We build our method upon the conditional image generation architecture SPADE~\cite{park2019SPADE}, and additionally address the problem of video consistency and background motion via a learned composition with optical flow based interpolation methods. The pose-guided neural rendering process is illustrated in Fig.~\ref{fig:generation}.

\paragraph{Foreground Human Image Generation.}

Our neural rendering model $G$ is built upon a conditional encoder-decoder architecture (see model details in the supplementary materials). The conditional image is first downsampled by an encoder, goes through a series of SPADE residual blocks~\cite{park2019SPADE}, and is finally upsampled by a decoder to generate the output image. In our case, the conditional image represents the 2D joints $\boldsymbol{p}_t$ and connecting limbs while the target image $\hat{F}_t$ depicts a person corresponding to the input pose. Motivated by \cite{wang2018vid2vid}, in addition to the skeletal image, we also input the image generated at the previous time step $\hat{I}_{t-1}$ to improve temporal consistency.

\paragraph{Adaptive Foreground-Background Composition.}
\label{sect:mask}
In real videos, backgrounds can also be dynamic, e.g., due to camera motion. These background motions, however, cannot be covered by the human motion model and are not reflected in the 2D skeleton maps. On the other hand, compared to human body motion, background motion is generally subtle and can be modelled by optical flow based interpolation techniques~\cite{bao2019depth,xu2019quadratic}. With the goal of preserving both foreground consistency and background dynamic, we propose to learn an adaptive foreground-background composition for neural rendering. 

We first generate an interpolated frame $\hat{B}$ with accurate background regions from the off-the-shelf video interpolation model (e.g., DAIN~\cite{bao2019depth} in our implementation) using the image pair $(I_{t-1},I_{t+1})$. Our neural rendering model would generate a foreground human image $\hat{F}$ alongside an alpha blending mask $\hat{M}$ by taking the skeleton $\boldsymbol{p}$ and the background $\hat{B}$ as inputs. Finally, the foreground and background images are composited according to the predicted alpha blending mask. The final output is obtained by 



\begin{equation}
\label{eqn:generation}
\begin{split}
F_t,\hat{M}_t = G(  \hat{B}_{t},\boldsymbol{p}_t,\boldsymbol{p}_{t-1},\hat{I}_{t-1}) \\
\hat{I}_t = \hat{F}_t \odot \hat{M}_t + \hat{B}_t \odot (1 - \hat{M}_t). 
\end{split}
\end{equation}
Here $\odot$ denotes the element-wise multiplication.

\begin{figure*}
\centering
\includegraphics[width=\linewidth]{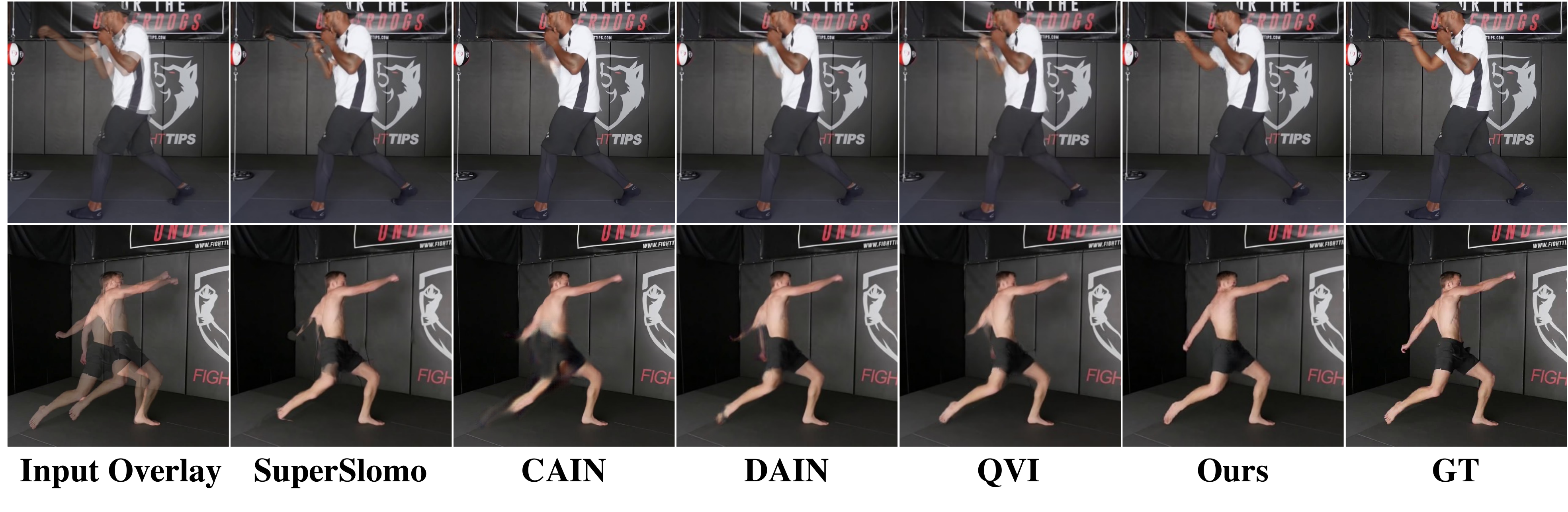}
\caption{\textbf{Qualitative comparison on HumanSloMo}. We generate intermediate frames from a low FPS video using our method and various video interpolation baselines. Our generated results are perceptually close to the ground truth. Despite the large motion, our method does not suffer from the common interpolation artifacts such as missing, ghosting or blurry limbs.}
\label{fig:comparison}
\end{figure*}

\paragraph{Training.}

At training time, we use only the low FPS video $ V_{low} = \left \{ I_0...,I_T \right \} $ and the corresponding 2D poses $\boldsymbol{P}_{low}= \{\boldsymbol{p}_0,...,\boldsymbol{p}_T \}$.
We train the network by reconstructing $I_t$ at each time step using the corresponding pose $\boldsymbol{p}_t$. For instance, when synthesizing the frame $I_t$ at the time step $t$, we precompute the background image $\hat{B}_t$ using the image pair $(I_{t-1},I_{t+1})$. Our network generate the foreground image $\hat{F}_t$, the alpha mask $\hat{M}_t$, and the composition image $\hat{I}_t$ via Equation~\ref{eqn:generation}. For each training iteration, we generate $K$ consecutive frames by iteratively providing the image generated at the previous time step (i.e., $\hat{I}_{t-1}$). Note that ground-truth image $I_{0}$ is provided instead for synthesizing the first frame at each iteration. To optimize the network parameters, we minimize the reconstruction loss and the perceptual loss \cite{Gatys_2016_CVPR} between the generated frames $\hat{I}_t$ and the ground-truth frames in the low FPS video ${I}_t$:
\begin{equation}
    \mathcal{L}_{image} = \sum_{t=1}^{T-1} (\Vert I_t - \hat{I}_t \Vert_{1} + \lambda_{percep} \Vert \psi(I_t)-\psi(\hat{I}_t)  \Vert_1),
\end{equation}
where $\psi_l$ is a trained deep feature extractor from the VGG-19 network~\cite{Simonyan15} pre-trained on ImageNet and $\lambda_{percep}$ is a scalar weighting the two terms. In addition we also encourage the generated foreground $\hat{F}_t$ to be similar to ${I}_t$ using the human-centric masks $M_t$ generated from ${P}_{low}$ following~\cite{ma2017pose}:
\begin{equation}
    \begin{split}
    \mathcal{L}_{fg} = \sum_{t=1}^{T-1} (\Vert (I_t - \hat{F}_t)\odot M_t \Vert_{1} +\\ \lambda_{percep} \Vert \psi(I_t\odot M_t)-\psi(\hat{F}_t\odot M_t)  \Vert_1).
    \end{split}
\end{equation}
Finally, to encourage the usage of $\hat{B}_t$ to handle the background and static part of human bodies, we regularize the predicted mask $\hat{M}_t$ with
\begin{equation}
    \mathcal{L}_{mask} = \sum_{t=1}^{T-1}\Vert \hat{M}_t \odot (1-M_t) \Vert_{1}.
\end{equation}
The overall loss function is the summation of the above terms plus an adversarial objective $\mathcal{L}_{adv}$ (see our supplementary materials):
\begin{equation}
    \label{eq:image}
    \mathcal{L}_{total} =\mathcal{L}_{adv} + \lambda_{im}\mathcal{L}_{image} + \lambda_{fg}\mathcal{L}_{fg} + \lambda_{m}\mathcal{L}_{mask},
\end{equation}
where $\lambda$s are weights that control the interaction of the loss.

\paragraph{Inference.}
The trained neural rendering model can synthesize intermediate frames $I_{\tau}$ not seen in the low FPS video $V_{low}$. To be specific, we first obtain the corresponding pose $\boldsymbol{p}_{\tau}$ following the motion modelling procedure described in Sec.~\ref{sect:motion}. We then interpolate the background image $\hat{B}_{\tau}$ from the two nearest frames to ${\tau}$ in the low FPS video. When generating the intermediate frame for 3-frame-triplet evaluation, the real image is provided as previous frame input. While when synthesizing multiple frames between two low FPS keyframes, the generated previous frames are provided iteratively. 
Finally, we obtain the desired frame $\hat{I}_{\tau}$ following Equation~\ref{eqn:generation}. 


\section{Experiments}

\subsection{Evaluation Protocol}
\label{sec:evaluation_protocol}

\paragraph{Dataset.} Existing video interpolation benchmarks focus on general scenes and are not adequate for our evaluation. We collect a new human action interpolation dataset \textbf{HumanSloMo} containing 80 sequences recorded at 30-240 FPS, capturing various high-speed  activities such as boxing and dancing performed by 10 subjects. For test, videos are downsampled to 15 FPS, from which high FPS videos are to be recovered and compared with the ground truth.

\paragraph{Baselines.} We compare our proposed method with state-of-the-art \emph{two-frame} video interpolation, including optical-flow based methods \textbf{SuperSlomo}~\cite{jiang2018super}, \textbf{DAIN}~\cite{bao2019depth}, \textbf{BMBC}~\cite{BMBC}, and implicit generation methods \textbf{CycleGen}~\cite{liu2019cyclicgen} and \textbf{CAIN}~\cite{choi2020cain}. We also evaluate recent \emph{four-frame} non-linear interpolation methods \textbf{QVI}~\cite{xu2019quadratic} based on PWC-Net~\cite{Sun2018PWC-Net} and HumanFlow~\cite{MultiHumanflow}.

\paragraph{Metrics.}
Following the protocol in~\cite{xue2019video,choi2020cain}, we generate one intermediate frame between two consecutive frames from the low FPS video and compute the following metrics:
\begin{itemize}[nosep]
    \item \textbf{PSNR} measures the ratio between the maximum possible power of a signal and the power of corrupting noise, indicating the per-pixel color difference.
    \item \textbf{LPIPS}~\cite{zhang2018perceptual} measures the difference between the feature maps extracted from the generation and the ground truth using a trained CNN.
    \item \textbf{SSIM}~\cite{wang2004image} measures the structural similarity between the generated image and the ground truth.
    \item \textbf{FVD}~\cite{unterthiner2019fvd} measures the overall visual quality and temporal coherence of the whole generated sequence, based on Fréchet Inception Distance~\cite{heusel2017gans}.
\end{itemize}
Since the human part in the image is of greater interest to us, we further report these metrics averaged over the human regions using the human-centric masks $M_t$, denoted by \textbf{mask PSNR/LPIPS/SSIM}.

For the distributional metric, we follow the evaluation protocol in~\cite{unterthiner2019fvd} by sampling the generated and ground-truth 21-frame sequences with a batch size of 32. We compute FVD between generated and ground-truth data in total 30 runs. The averaged value of 30 runs is reported.
\begin{table*}[t]
    \centering
\begin{tabular}{lcccc}
\toprule
Methods              &    PSNR / mask PSNR($\uparrow$) & SSIM / mask SSIM($\uparrow$)  & LPIPS / mask LPIPS($\downarrow$) & FVD($\downarrow$)    \\ 
\midrule
 Vid2vid~\cite{wang2018vid2vid}  
                     & 21.01 / 20.17 & 0.7619 / 0.9467 & 0.1967 / 0.0352 & 309.60  \\
 PATN~\cite{zhu2019progressive}
                     & 26.22 / 20.52 & 0.8956 / 0.9511 & 0.1110 / 0.0365 & 319.75  \\
\midrule
 CyclicGen~\cite{liu2019cyclicgen}
                     & 28.90 / 22.87 & 0.9502 / 0.9609 & 0.0529 / 0.0315 & 203.02 \\
 SuperSlomo~\cite{jiang2018super}
                     & 29.39 / 23.47 & 0.9556 / 0.9669 & 0.0466 / 0.0256 & 190.58 \\
 BMBC~\cite{BMBC} &  29.80 / 23.83 & 0.9617 / 0.9712 & 0.0449 / 0.0239 &  196.41 \\

 CAIN~\cite{choi2020cain}         
                     & \underline{30.52} / \underline{24.59} & 0.9642 / 0.9731 & 0.0419 / 0.0251 & \underline{157.81} \\
 DAIN~\cite{bao2019depth}                 
                & 30.42 / 24.50 & \textbf{0.9655} / \underline{0.9740} & \underline{0.0400} / \underline{0.0222} & 160.23 \\

 Ours           & \textbf{30.75} / \textbf{24.93} & \underline{0.9648} / \textbf{0.9745} & \textbf{0.0395} / \textbf{0.0200} & \textbf{123.04} \\
\midrule
 QVI-HumanFlow~\cite{MultiHumanflow}                 
                & 29.63 / 23.78 & 0.9604 / 0.9708 & 0.0484 / 0.0262 & 209.15 \\   
 QVI-PWCNet~\cite{xu2019quadratic}                 
                & 30.75 / 25.01 & 0.9657 / 0.9759 & 0.0415 / 0.0211 & 139.35 \\
 Ours$\dagger$           & \textbf{31.00} / \textbf{25.36} & \textbf{0.9658} / \textbf{0.9773} & \textbf{0.0409} / \textbf{0.0198} & \textbf{121.86} \\
\midrule
\midrule
 Ours w/o fusion     & 26.83 / 23.46 & 0.8873 /	0.9668 & 0.1696 / 0.0278 & 219.38 \\
 Ours w/o prev. frame & 30.16 / 24.39 & 0.9631 / 0.9734 & 0.0418 / 0.0207 & 129.91 \\
 Ours w. linear pose  & 30.22 / 24.58 &	0.9627 / 0.9734 & 0.0435 / 0.0215 & 143.45 \\
 Ours w. pred. pose    & 30.75 / 24.93 & 0.9648 / 0.9745 & 0.0395 / 0.0200 & 123.04 \\
 Ours w. GT pose      & \textbf{31.37} / \textbf{25.60} & \textbf{0.9665} / \textbf{0.9761} & \textbf{0.0384}	/ \textbf{0.0192} & \textbf{109.73} \\

\bottomrule

\end{tabular}
\vspace{0.5em}
\caption{\textbf{Quantitative comparison on HumanSloMo.} We report the PSNR, SSIM, LPIPS and FVD wrt. the ground truth frames for various video interpolation baselines and our methods. Please refer to Sec.~\ref{sec:evaluation_protocol} for details regarding metrics and baselines.}

\label{tab:exp1}
\end{table*}

\begin{table}[t]
\resizebox{\linewidth}{!}{%
\centering
\setlength{\tabcolsep}{0.1cm}
\begin{tabular}{lccc|cccc}
\toprule
         & \begin{tabular}[c]{@{}c@{}}1D\\ CNN\end{tabular}  & \begin{tabular}[c]{@{}c@{}}2D\\ CNN\end{tabular} & Seq2seq & Linear & \begin{tabular}[c]{@{}c@{}}Ours w/o\\ Denoising\end{tabular} &  \begin{tabular}[c]{@{}c@{}} Ours$\ddagger$\\ Quadratic\end{tabular}& Ours  \\
\midrule
 avg L1 ($\downarrow$)         & 3.194  & 3.260  & 1.170  & 3.767  & 0.917   &  1.870 & \textbf{0.820}  \\
 max L1 ($\downarrow$)         & 80.543 & 75.774 & 28.818  & 96.838& 23.467   & 26.958 & \textbf{19.581} \\
\bottomrule
\end{tabular}
}
\vspace{0.1em}
\caption{\textbf{Ablation study on human motion modelling}. We evaluate ablative baselines of human motion modelling on 2D motion sequences simulated from AMASS. We measure the mean and max L1 error among all joints. }
\label{tab:exp2}
\end{table}


\subsection{Video Interpolation Comparison to SOTA }
\label{sec:comparison_sota}

We first conduct the comparison with other SOTA video interpolation methods. Fig.~\ref{fig:comparison} indicates that our generated results are perceptually closer to the ground truth than other baseline methods. It can be seen that the baselines struggle to model large motion between input frames and produce artifacts such as ghosting, missing or blurry limbs, while our method generates complete and realistic human bodies. Moreover, our generated body poses align significantly better with the ground-truth images. 

Table~\ref{tab:exp1} summarizes the quantitative results. Our method achieves SOTA performance on pixel-level metrics. When considering only the area of interest, indicated by the ground-truth foreground mask, our method outperforms other baselines consistently (cf. mask PSNR/SSIM/LPIPS). While non-linear methods \textbf{QVI} slightly improve the metrics using more information for synthesizing the frame, it still suffers from large displacements of fast human motion in Fig.~\ref{fig:comparison}. Moreover, our pipeline can further improve the non-linear model by using its background information (denoted as \textbf{Ours$\dagger$}).

Pixel-level metrics do not always reflect the perceptual plausibility since the task of motion interpolation is inherently uncertain and stochastic. That is, there might be several feasible joint trajectories, which do not fully agree with the ground truth. Thus we employ the distributional metric FVD to better quantify visual quality and temporal coherence, where ours outperforms existing ones with a noticeable margin, indicating improved realism and visual quality.


\subsection{Ablation Study}
\label{sec:ablation_motion}

We conduct controlled experiments to verify the effectiveness of our design choices in both human motion modelling module and neural rendering module.

\paragraph{Human Motion Modelling.}
We generate noisy 2D poses in 7.5 FPS for test following the data generation procedure described in Sec.~\ref{sec:motion_training} from 500 held-out motion sequences in AMASS. We measure the average L1 difference between recovered 60 FPS motion sequence and the ground truth, denoted by \textbf{avg L1}. Moreover, since extreme failure cases with large pose errors could severely affect the quality of the generated image, we also report the average of \emph{maximum} L1 error among all joints, denoted by (\textbf{max L1}).

Results are summarized in Table~\ref{tab:exp2}. First of all, simple linear interpolation (\textbf{Linear}) leads to unsatisfactory results due to the non-linear dynamics of human motion and the noise in the input poses. By introducing our motion interpolation network \textbf{Ours w/o denoising}, the pose error is significantly reduced (3.767 vs 0.917). With the additional denoising network, our full motion modelling pipeline \textbf{Ours} is complete and achieves the lowest pose error. Alternative backbones are also evaluated, including \textbf{1D CNNs}~\cite{aberman2019learning,transmomo2020}, \textbf{2D CNN}~\cite{kaufmann2020infilling} and \textbf{Seq2Seq}~\cite{harvey2018recurrent}, demonstrating that transformer-based architecture indeed leads to the best performance. We also replace linear interpolation by a higher order one in our model (denoted as \textbf{Ours$\ddagger$}). Since our transformer model is capable of predicting non-linear motion using attention mechanism, quadratic interpolation does not contribute improvement.

\paragraph{Pose-Guided Neural Rendering.}

\begin{figure}
\centering
\includegraphics[width=\linewidth]{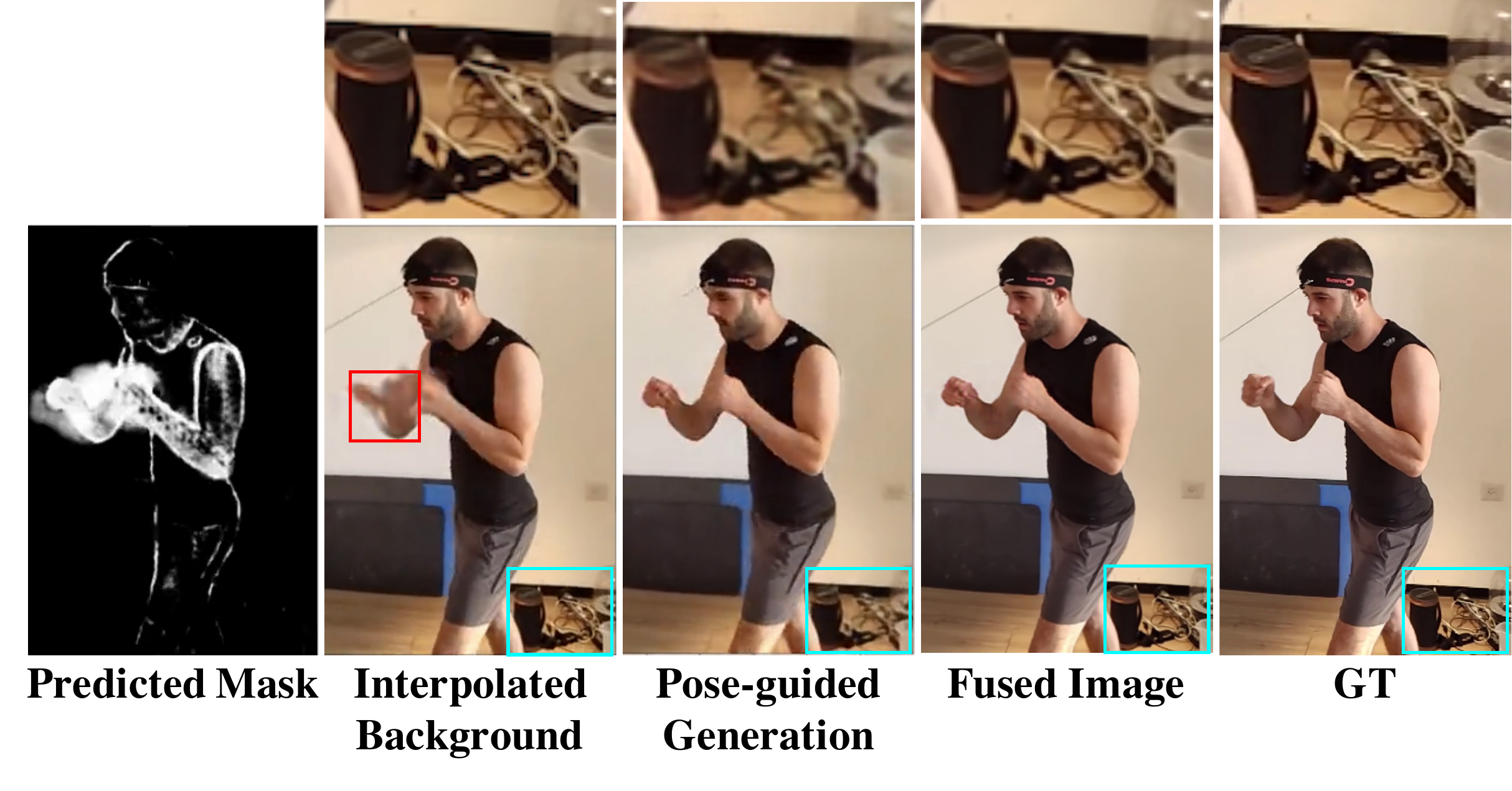}
\caption{\textbf{Qualitative result of foreground and background fusion.} Our predicted mask assigns more weight to the foreground human image generated by our network for composition. This weight correlates well to the actual quality of corresponding parts and therefore the final output combines the best of both images. }
\label{fig:exp_fusion}
\end{figure}

\begin{figure}
\centering
\includegraphics[width=\linewidth]{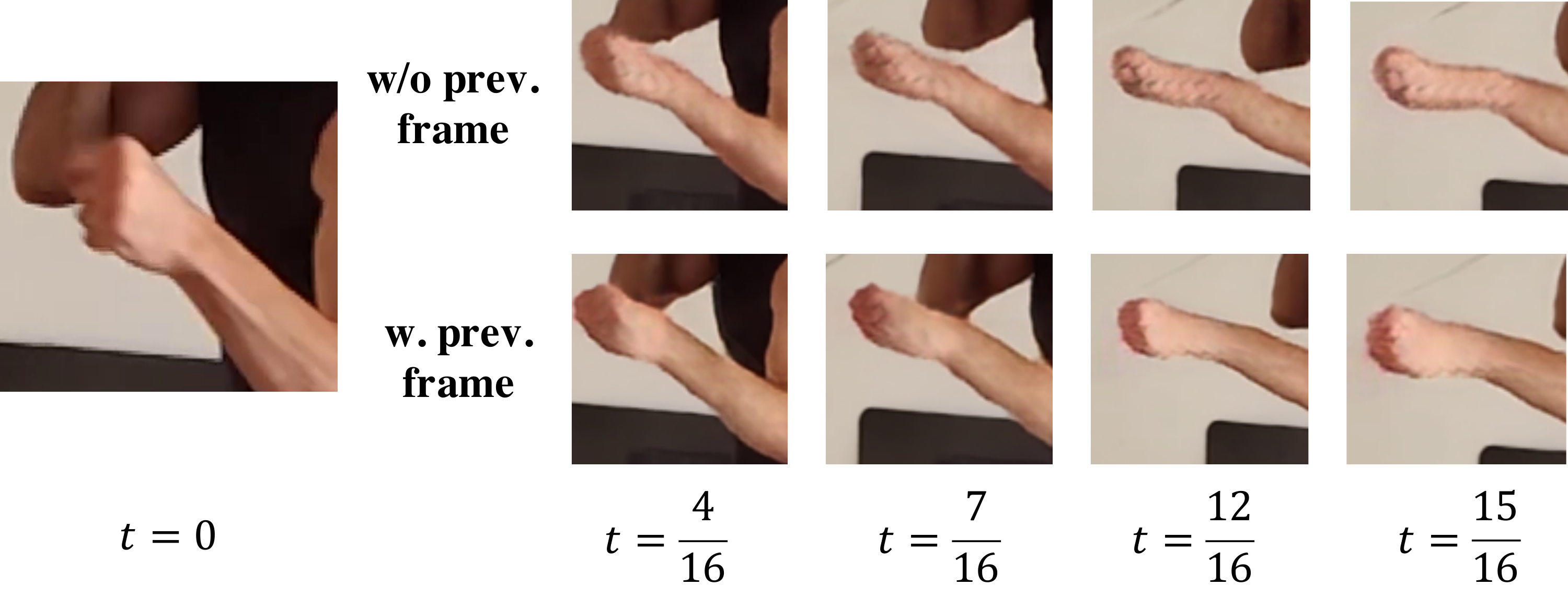}
\caption{\textbf{Qualitative result with and without using previous frames.} Video sequences generated with and without iteratively taking the previous generation as input. With the previous information, the fist's shape is preserved better and more consistently over time. }
\label{fig:exp_autoregressive}
\end{figure}

As discussed in Sec.~\ref{sect:mask}, the neural rendering model is not designed to infer background pixels and we propose to learn the composition of foreground and background images instead. As shown in Table~\ref{tab:exp1}, without fusion (\textbf{Ours w/o fusion}), the performance (e.g. PSNR) decreases significantly due to inconsistent backgrounds. As shown in Fig.~\ref{fig:exp_fusion}, pose-guided neural rendering performs well on the human region while the optical flow based interpolation method \cite{bao2019depth} is more stable on the background and static regions. The alpha mask is correctly predicted to fuse both images. We also find that providing previously generated frames as conditional inputs can improve temporal consistency and reduce the shape distortion, as shown in Fig.~\ref{fig:exp_autoregressive}. Removing this feature reduces the accuracy as reported in Table~\ref{tab:exp1} (\textbf{Ours w/o prev. frame}). 

We also compare the neural rendering module with methods that are capable of generating human images based on poses, including \textbf{Vid2vid}~\cite{wang2018vid2vid} and \textbf{PATN}~\cite{zhu2019progressive}. We train their networks with the same data as our method requires, namely the low FPS videos and the corresponding 2D poses. As shown in Table~\ref{tab:exp1}, without our proposed adaptations these methods do not yield comparable results.

\paragraph{Human Motion Modelling for Video Interpolation.}
To understand how important human motion modelling indeed is, we generate intermediate frames using linearly interpolated joints. As shown in Table~\ref{tab:exp1} \textbf{Ours w. linear pose}, this leads to a performance decrease especially in terms of realism and visual quality measured by FVD, showing that modelling the non-linear dynamics of the human motion is crucial for video interpolation.

We also evaluate the performance of our method given known human motion. We use 2D poses detected from the ground-truth images to generate these intermediate frames. As shown in Table~\ref{tab:exp1} as \textbf{Ours w. GT pose}, significant improvements are gained across all metrics, showing that the motion interpolation task contains inherent uncertainty. 



\begin{figure}
\centering
\includegraphics[width=\linewidth]{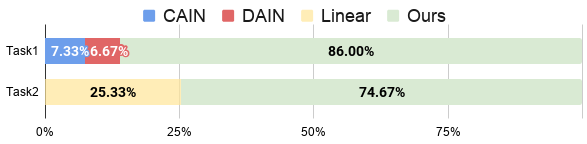}
\caption{\textbf{Result of user studies.} The participants voted the best result for each instance. Our generated high FPS sequences are consistently preferred in both tasks.}
\label{fig:bar}
\end{figure}

\subsection{User Study}
Finally, we conducted two user studies to verify our proposed pipeline :
\begin{itemize}[nosep]
\item Task 1: Users are asked to select the preferred results among \textbf{CAIN}, \textbf{DAIN} and \textbf{Ours}.
\item Task 2: Users are asked to choose the preferred results between \textbf{Ours} and \textbf{linearly interpolated ones}. 
\end{itemize}
For each task, 30 participants are invited to make decisions for 10 different sequences. As shown in Fig.~\ref{fig:bar}, 7.33\% choose CAIN, 6.67\% choose DAIN, 86.0\% choose Ours. Moreover, Ours is preferred 74.67\% when compared with linearly interpolated motion. The results confirm our findings from Sec.~\ref{sec:comparison_sota} and Sec.~\ref{sec:ablation_motion}.



\section{Conclusion}

In this paper, we propose a novel method for the synthesis of videos of human activities. Our pipeline uniquely integrates modules of human motion modelling and pose-guided neural rendering, which enables the reasoning about challenging articulated motion with highly non-linear dynamics. We believe that the further cross-pollination of ideas from these two rapidly developing areas is a exciting and up-and-coming direction for video synthesis.

\paragraph{Acknowledgements}Xu Chen was supported by the Max Planck ETH Center for Learning Systems.




{\small
\bibliographystyle{ieee_fullname}
\bibliography{egbib}
}
\end{document}